\documentclass{article}

\usepackage{graphicx}
\usepackage{amsmath}
\usepackage{amssymb}
\usepackage[margin=3cm]{geometry}
\usepackage{setspace}

\usepackage{natbib}

\usepackage{inputenc}
\usepackage{bm}
\usepackage{booktabs}
\usepackage{microtype} 
\usepackage{moreverb}

\usepackage{hyperref}

\title{Deep learning the Hurst parameter of linear fractional processes and assessing its reliability}

\date{December 1, 2023}

\author{D\'aniel Boros\thanks{Department of Probability Theory and Statistics, E\"otv\"os Lor\'and University, Budapest, Hungary} \and B\'alint Csan\'ady \thanks{Institute of Mathematics, E\"otv\"os Lor\'and University, Budapest, Hungary} \and Iv\'an Ivkovic  \thanks{Alfr\'ed R\'enyi Institute of Mathematics, Budapest, Hungary} \and L\'or\'ant Nagy  \thanks{Alfr\'ed R\'enyi Institute of Mathematics, Budapest, Hungary} \and Andr\'as Luk\'acs  \thanks{Institute of Mathematics, E\"otv\"os Lor\'and University, Budapest, Hungary} \and L\'aszl\'o M\'arkus  \thanks{Institute of Mathematics, E\"otv\"os Lor\'and University, Budapest, Hungary; Department of Statistics, University of Connecticut, Connecticut, USA} }

\begin{document}

\maketitle

\begin{abstract}
This research explores the reliability of deep learning, specifically Long Short-Term Memory (LSTM) networks, for estimating the Hurst parameter in fractional stochastic processes. The study focuses on three types of processes: fractional Brownian motion (fBm), fractional Ornstein-Uhlenbeck (fOU) process, and linear fractional stable motions (lfsm). The work involves a fast generation of extensive datasets for fBm and fOU to train the LSTM network on a large volume of data in a feasible time. The study analyses the accuracy of the LSTM network's Hurst parameter estimation regarding various performance measures like RMSE, MAE, MRE, and quantiles of the absolute and relative errors. It finds that LSTM outperforms the traditional statistical methods in the case of fBm and fOU processes; however, it has limited accuracy on lfsm processes. The research also delves into the implications of training length and valuation sequence length on the LSTM's performance. The methodology is applied by estimating the Hurst parameter in Li-ion battery degradation data and obtaining confidence bounds for the estimation. The study concludes that while deep learning methods show promise in parameter estimation of fractional processes, their effectiveness is contingent on the process type and the quality of training data.
\end{abstract}


\vspace{3cm}

\pagebreak

\section{Introduction}

Fractional processes play a pivotal role in the stochastic modeling of diverse phenomena, including the wear and tear of machinery and the valuation of financial instruments. A key variable in these models is the Hurst exponent, which may reflect memory decay, self-similarity, fractal dimension, or all of them simultaneously, as in fractional Brownian motion.

The model's accuracy is highly sensitive to the Hurst exponent, highlighting its criticality; even slight errors in estimating this parameter can cause significant discrepancies in the model's effectiveness and predictive capability. Consequently, it is vital to comprehend and precisely quantify the margin of error associated with this parameter's estimation.
While conventional statistical methods for estimating this parameter have been thoroughly explored, the potential of deep learning techniques in this area warrants investigation. Our research focuses on three process types: fractional Brownian motion (fBm), fractional Ornstein-Uhlenbeck (fOU) process, and fractional Lévy stable motions (fLsm). \\

Degradation modeling and reliability analysis are fundamental to the Prognostics and Health Management (PHM) of contemporary complex systems. With advancements in measurement technologies, a notable long-range dependence (LRD) has been observed in the degradation patterns of various assets, including turbofan engines, blast furnaces, lithium-ion batteries, and chemical catalysts. This LRD, also known as long-term memory effect or persistence, indicates that correlations of asset degradation increments extend over extensive periods. Therefore, future degradation trends are not only influenced by the current condition of the asset but are also deeply linked to its entire degradation history. The LRD's significance extends beyond asset performance degradation, being prevalent in fields like finance, hydrology, and biology. Consequently, numerous models integrating LRD  have been developed for asset degradation and reliability analysis.

To monitor the long-term memory LRD in degradation patterns effectively, several researchers have incorporated fBm in asset degradation models. The memory structure of FBM is delineated by the Hurst parameter $H$, which lies within the range $0 < H < 1$. Specifically, when $0.5 < H < 1$ fBm exhibits LRD, making it increasingly relevant for analyzing LRD-integrated degradation data across various assets. Examples of its application include Xi et al.'s \cite{XiChenZhou2017} model that leverages the LRD for predicting the remaining useful life of turbo engines, Zhang et al.'s \cite{ZhangZhouChenShang2019} integration of multiple degradation modes in an fBm-based model, Si et al.'s \cite{SiShaoWei2020} application in accelerated degradation tests, and Zhang et al.'s \cite{ZhangMoWangMiao2020} development of an explicit probability density function (PDF) for remaining useful life estimation within an LRD-integrated framework. These models exemplify the growing scope of fBm applications in asset degradation analysis \cite{ZhangZhouChenXi2019,SongChenCattaniZio2020,WangSongZioKudreykoZhang2020}. The fBm solely relies on the Hurst parameter to characterize the degradation process, which may not sufficiently capture complex time series data attributes. Nonetheless, practical degradation processes typically exhibit non-Gaussian traits, whereas the fBm process conforms to Gaussian distributions.

Conversely, Lévy stable motion exhibits non-Gaussian stable distributions and is characterized by parameters $\alpha$, $\beta$, $\gamma$, and $\delta$. This motion, simplifying to Brownian motion when $\alpha = 2$, is distinguished by its heavy-tailed nature, with a probability density that diminishes following a power law. The fractional Lévy stable motion (fLsm), as an extension of Lévy stable motion, encapsulates both non-Gaussian characteristics and heavy-tailed properties, as well as exhibiting LRD. In this model, positive LRD is indicated when $H > \frac{1}{\alpha}$, suggesting continuity in future trends with the present. Conversely, $H < \frac{1}{\alpha}$ denotes the absence of LRD, implying contrasting future trends. When $H = \frac{1}{\alpha}$, the process is deemed independent of past trends. Notably, in the $\alpha = 2$ case, the fLsm model aligns with the fBm. Thus, fLsm presents a versatile framework capable of modeling a broad spectrum of stochastic processes, ranging from heavy-tailed to both Gaussian and non-Gaussian distributions.

In the realm of industrial production, where sensor data is abundantly collected, data-driven methodologies are increasingly vital for predicting performance degradation in complex mechanical systems. These methods are broadly categorized into statistical and deep learning-based approaches. Deep learning techniques, including convolutional neural networks (CNN), long short-term memory (LSTM), deep belief networks (DBN), deep autoencoders (DAE), and transfer learning (TL), have garnered significant attention in RUL prediction. For instance, Ding et al. \cite{DingYangChengYang2021} introduced an RUL prediction method for rolling bearings based on DCNN, enhancing feature learning capabilities. Qin et al. \cite{QinChenXiangZhu2020} developed a neural network with a gated dual attention unit for predicting the RUL of rolling element bearings. Yan et al. \cite{YanQinXiangWangChen2020} crafted an ON-LSTM network for gear RUL estimation, integrating tree structures within LSTM to improve prediction accuracy. Haris et al. \cite{HarisHasanQin2021} combined DBN with Bayesian optimization for estimating supercapacitor RUL, while Wang et al. \cite{WangPengLiuXuLiuSaeed2020} utilized DAE for feature extraction and LSTM for electric valve RUL prediction. Fan et al. \cite{FanNowaczykRognvaldsson2020} proposed a feature-based TL approach to extend RUL prediction models from simpler to more complex domains.

While deep learning approaches have achieved notable successes in predicting the remaining useful life (RUL) of equipment, these algorithms' effectiveness is heavily reliant on the quality and volume of the training data, leading to inefficiencies in model training, particularly with combinatorial network models. Additionally, the determination of hyper-parameters in these algorithms often involves laborious cross-validation or heuristic methods, limiting their practical application in industrial settings. A significant limitation of most deep learning methods is their focus on point-wise RUL predictions, which struggle to accurately quantify the uncertainty in RUL projections under varying conditions, thereby inadequately addressing risk assessment in predictions.
In contrast, statistical data-driven methods, which do not depend on extensive degradation data, possess intrinsic strengths in accurately quantifying the uncertainties associated with RUL predictions, offering a more robust framework for risk assessment.
\ \\

The primary aim of our research is the swift and accurate estimation or calibration of these fractional processes in complex models used across various sectors, such as finance and reliability engineering. In the domain of financial volatility modeling, the Hurst exponent frequently falls below 0.5, leading to rough paths. Conversely, in reliability engineering, particularly in degradation or remaining useful life (RUL) analysis, the Hurst exponent often exceeds 0.5, indicating a long memory effect. 

We train neural networks using extensive and accurately simulated datasets, ensuring the Hurst exponent's variation covers its entire spectrum from 0 to 1. To generate our training samples, we implement a sophisticated Davies-Harte-type algorithm capable of efficiently producing sample paths from isonormal processes, encompassing the fBm and fOU processes. For estimating parameters of the fOU process, we utilize a bidirectional Long Short-Term Memory (LSTM) network, comprising two layers and enhanced with a normalization layer.

While current deep-learning-based research predominantly concentrates on \textit{predicting} risks or losses, the dynamics of risk propagation and methods to impede this spread are comparatively underexplored. To enhance production safety, product reliability, or financial stability, a blend of targeted risk response strategies and an ambient, all-encompassing risk prevention framework is essential. In addressing this, we propose integrating the study of \textit{propagation dynamics} with deep learning methodologies to thoroughly model and scrutinize the spread of risks. Keeping this perspective, our goal is to utilize deep learning to unravel the dynamic parameters of stochastic processes that underlie or stimulate risk propagation.

In recent years, a number of works utilizing neural networks emerged on the estimation of the Hurst parameter of fBM.
Those applying  multilayer perceptions (MLPs) have to cope with the required fixed input size and hence have one of the following alternatives: 
either inference can be performed only on a fixed-length series \citep{ledesma11,han20}, or inference is made on a set of process-specific statistical measures enabling a fixed size input to the neural networks \citep{kirichenko22,mukherjee23}.
A more recent signature-based method, described in Bonnier et al. (2019) \cite{Bonni19}, is also capable of estimating fBM's Hurst parameter. In executing that, the extracted statistical descriptors are processed by an LSTM.

The hybrid application of statistical descriptors and neural networks employed in the mentioned methods does not bring significant improvement compared to our purely neural network solution while increasing the computational time sometimes unbearably.
Another common weakness of the recently published methods is that they do not address the possible limitations caused by scaled inputs.

A major advancement of our research lies in the creation of a neural network-based methodology, which markedly surpasses traditional statistical methods in determining the parameters of the fBm and fOU processes. This method excels in accuracy and speed compared to previous statistical and deep learning approaches. To achieve it, we need large-scale teaching of the network, and to do that, we have successfully invented a very fast generation of extensive fBm datasets for training purposes.\\
\ \\

   

\section{Teaching the Neural Network} 

\subsection{The Processes}

A \textbf{fractional Brownian motion (\textbf{fBm})} \cite{Mandel68}, denoted as $B_{H}(t)$, is a Gaussian process initiating from zero and characterized by continuous time, zero mean, and the autocovariance function 
$$E\left[ B_{H}(t) \cdot B_{H}(s)\right] = \frac{1}{2}\left( |t|^{2H} + |s|^{2H} - |t-s|^{2H}\right).$$
The fBm is notable for having stationary and \textbf{dependent} increments and is recognized for being a \textbf{self-similar} process with \textbf{fractal} paths. The Hurst exponent $H$ is its sole parameter. According to Robert Adler's seminal work, \textbf{Brownian motion} and Stochastic Differential Equation (SDE) driven \textbf{diffusion processes} invariably exhibit fractal dimensions (FD) of 1.5. That highlights the inadequacy of these processes for modeling phenomena with varying FDs, thus underscoring the significance of fBm. Notably, the FD of fBm paths varies with the Hurst parameter $H$, so that FD$=2-H$.

A \textbf{fractional Ornstein-Uhlenbeck (fOU)} process is defined \cite{Cher2003} by a fractional stochastic Langevin differential equation 
 \begin{align}\label{fOU}
 dX(t) &= \kappa(\theta - X(t))dt + \sigma dB_H(t),
 \end{align}
where the process is driven by an fBm $B_{H}(t)$ with \textbf{Hurst} parameter $H \in (0, 1)$. Both the \textbf{drift} parameter $\kappa$ and the \textbf{volatility} parameter $\sigma$ are positive real constants. The solution to the fOU equation is known to exist and is unique, subject to an initial condition. It can be expressed explicitly, particularly for a 0 initialization and expectation, as follows:
	$$X(t) = - \sigma \int_0^t \text{e}^{-\kappa(t-s)} dB_H(s).$$
The fOU process is an \textbf{isonormal} and, therefore, a Gaussian process belonging to the first Wiener-Ito chaos driven by fBm. It achieves a unique stationary solution when initiated in a stationary state. The paths of the fOU process inherit their \textbf{fractal dimension} from the driving fBm. The fOU process is characterized by four parameters: $H$, $\kappa$, $\theta$, and $\sigma$. \cite{Biag2008} \cite{Cher2003}\cite{Cou2007}\\
\ \\

The fBm has different extensions to the $\alpha$-stable case. One of the most commonly used is the \textbf{linear fractional stable motion (lfsm)}. This process is also called linear fractional Levy motion (lfLm) or fractional L\'evy stable motion (fLsm).

    The lfsm $L^{\alpha}_H(t)$ is defined as the stochastic process given by the integral
    $$L^{\alpha}_H(t) = \int_{-\infty}^{\infty}  \left({\left({(t-x)_+}\right)^{H-1/\alpha} - \left({(-x)_+}\right)^{H-1/\alpha}}\right) dM(x)$$
    where $0 < \alpha < 2$, is the parameter of stability, $0<H<1$, $H \neq \frac{1}{\alpha}$ is the Hurst exponent and $M$ is an $\alpha$-stable random measure on $\mathbb{R}$ with Lebesgue control measure. For any $z\in \mathbb{R}$ $(z)_+ = max(z,0)$. 
    
 The lfsm is a fractional \textbf{self-similar} stable process with \textbf{stationary increments} (H-sssi), where $H$ is the Hurst exponent of fractionality ( for more details, see Samorodnitsky and Taqqu (1994) \cite{ST94}). The parameter $H$ characterizes the self-similarity property of lfsm. The marginal distribution of $L^{\alpha}_H(t)$ is $\alpha$ stable. It follows that
$P (L^{\alpha}_H(t) > x) \propto  t^{\alpha H} x^{-\alpha}$ as $x \to \infty$, and that the generalized dispersion defined in the quantile sense satisfies $D(L^{\alpha}_H(t)) \propto t^H,$  which is similar to the
behavior of fBm. If $1/\alpha < H < 1$, and $\alpha \in(1, 2)$, this process may be shown to present
long-range dependence in some extended sense given in Samorodnitsky and Taqqu (1994) \cite{ST94}. It has to be clearly differentiated from the stable time-changed Brownian motion described in Huillet \cite{Huill99}.\\
Let us recall that the path properties of a linear fractional stable motion strongly depend on the interplay between the parameters H and $\alpha$. When $H > 1/\alpha $, the lfsm paths are H\"older continuous on compact intervals of any order smaller than  $H - 1/\alpha$.
If  $H < 1/\alpha,$ the lfsm explodes at jump times of the driving L\'evy process; in particular, it has unbounded paths on compact intervals. 
Clearly, in this approach, the stable
character of the resulting process has been favored, and this model is the natural extension of the fBm in this respect. \\

\subsection{The neural network structure}
There are three kinds of invariances that we might require from the network: shift, scale, and drift invariance.
In order to make an fBm Hurst-estimator that works well in practice, we want to rely on all three of the above invariances.
We can obtain shift invariance by transforming the input sequence to the sequence of its increments.
Differentiating the input also turns drift invariance to shift invariance.
By performing a standardization on the sequence of increments, we can ensure drift and scale invariance.
The standardizing phase can also be considered as an additional layer to the network, applying the transformation $x \mapsto (x - \overline{x})/{\hat{\sigma}(x)}$ to each sequence of increments $x$ in the batch separately, where $\hat{\sigma}(x)$ is the sample standard deviation over the sequence $x$,
and $\overline{x}$ is the arithmetic mean over $x$. Note that the sample standard deviation is a biased estimator of the true one because of the autocorrelated samples. The bias, however, is negligible because of the relatively long sample length.


After the layers to ensure invariances, the next layers constitute a sequential regressor.
This part of the network first transforms the input sequence into a higher dimensional sequence, after which each dimension of the output is averaged out, resulting a vector.
Finally, a scalar output is obtained by an MLP.
In the present analysis, we only consider an LSTM network \citep{Hochreiter1997}.
We found that the specific hyperparameter configuration does not have a significant effect on its performance.
The following are the hyperparameters that we used in the experiments below.

The applied network consists of an unidirectional LSTM with two layers; its input dimension is 1, and the dimension of its inner representation is 128. In both models, we use an MLP of 3 layers (output dimensions of 128, 64, and 1), with a PReLU activation function between the first two layers. 
AdamW optimization on the MSE loss function was used for training the models \citep{Loshchilov2017}.

\subsection{The training procedure}

When faced with limited real data, the use of synthetic data generators enables the training of neural networks on virtually unlimited datasets. In this approach, the loss calculated on new data batches serves as the validation loss, as each batch consists entirely of new, synthetically generated data. This method helps prevent overfitting by ensuring the model does not specialize too heavily on the training data and also highlights the critical importance of the quality of the synthetic data generator.

A \textbf{large and high-quality simulated sample} from both fBm and fOU processes is applied for effective training of the neural network. We utilized Kroese's method \cite{kroe2013}, a variation of the Davies-Harte procedure, to generate fBm and fOU trajectories. The Hurst parameter for these trajectories is set to be uniformly distributed between 0 and 1. 

To ensure the quality of the simulation, we assess Gaussianity, compare the empirical autocovariance to the theoretical model, and re-estimate the parameters using classical statistical estimators, such as the R/S, Whittle, variogram, Higuchi, and Peng's detrended fluctuation analysis. While the original Davies-Harte method for generation is accessible in existing Python packages, our application of Kroese's method necessitates a custom \textbf{implementation} using efficient Python framework tools. Our implementation strategy preserves the covariance structure, saving it to avoid redundant computations, thereby significantly enhancing computational speed. As the lfsm process does not belong to the class of isonormal processes, the generation has to rely on different principles. In this case, we were unsuccessful in developing a similarly efficient generator; therefore, we cannot train the network on lfsm trajectories. Instead, we use the package rlfsm in R to generate lfsm sample paths. Simulation of the sample paths is done via Riemann-sum approximations of its symmetric $\alpha$-stable stochastic integral representation while
Riemann sums are computed efficiently using the Fast Fourier Transform algorithm. However, it is much slower, so creating a proper training set is not feasible this way. Therefore, we had to contend with analyzing the estimation of the Hurst parameter of lfsm processes by the LSTM network trained on fBm or fOU processes.

\textbf{Speed} is a critical factor in this context, as we employ up to $10^7$ trajectories of the length of 1600 or 6400 for training the network, meaning 16-64 billion data.

The neural network training is carefully designed to accommodate this complex setup. The model utilizes a learning rate of 0.0001 with the AdamW optimizer, a combination chosen for its precision in model tuning and effective convergence. For loss functions, both L1Loss (Mean Absolute Error) and MSE (Mean Squared Error) are employed, crucial in refining the model's prediction accuracy by appropriately weighting different types of errors. The training spans various epoch lengths – 25, 50, and 100 epochs – with each epoch generating 100,000 sequences. This strategy allows the model to learn at multiple depths and assess the impact of different training durations. Moreover, the batch sizes are set at 32 for training and 128 for validation, ensuring efficient learning with smaller batches during training and a broader assessment of data during validation. We execute several trainings with trajectory lengths of 400, 1600, and 6400 and analyze the obtained estimator.

The overall success of this method depends significantly on the generator's accuracy and reliability. If the generator fails to closely approximate the desired distribution, there's a risk that the model might learn the generator's errors, leading to distorted results.

\section{Analysis of the Hurst parameter estimation}
\subsection{Evaluation of the LSTM estimator on fBm and fOU samples}
For benchmarking, we choose the estimator obtained by training the network on fBm trajectories of length 1600 with a training span of 100 epochs. Increasing the number of epochs or the length of the training series does not result in much decrease in the loss function; hence, the choice. We choose the mean squared error (MSE) and the mean absolute error (MAE) for the loss function. The (root)(R)MSE is minimized by the conditional mean, making it most useful when large errors are particularly undesirable. Given the constraint on the estimated values, large errors are of less concern in this case. The MAE is a linear score, which means that all the individual differences are weighted equally in the average. Minimizing MAE will make the fit closer to the median and eventually more biased. 

The trained network is then used to estimate the Hurst parameter of 10000 fBm and fOU trajectories of length 400, 1600, and 6400, respectively. We evaluate the estimation in terms of the RMSE, the MAE, and the mean relative errors (MRE). The MAE and the RMSE can be used together to diagnose the variation in the errors in a set of predictions. The RMSE will always be larger or equal to the MAE; the greater the difference between them, the greater the variance in the individual errors in the sample. The Hurst parameter is an exponent; therefore, the absolute or the relative deviation will heavily affect the modeling accuracy, necessitating their analysis. It is not just the overall variability of the errors that reflects the loss or risk in the model. Rarely occurring severe errors can be very much intolerable in some applications. Therefore, beyond the mean error values, we also calculate the 95\% quantiles and the medians of the absolute and relative errors together with the maximum of the absolute error. The reason is that in real applications, we estimate the Hurst exponent trajectory-wise, and even if the estimation's significant inaccuracy occurs only in a small percentage of the observations (i.e., stock prices, degradation of sensitive machine parts, etc.), it still can cause unbearable loss or risk.

\begin{figure}
    \centering
    \includegraphics[width=0.90\textwidth]{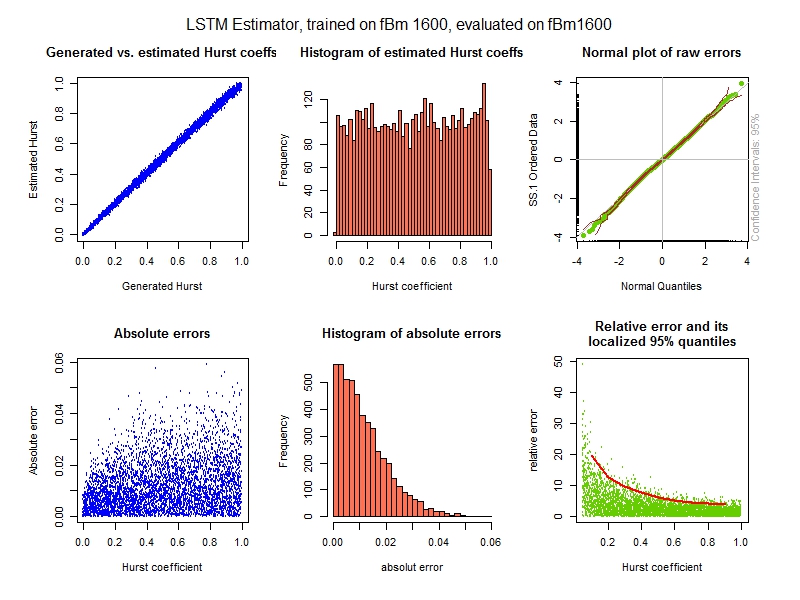}
    \caption{LSTM estimation and its errors. The network is trained on 1600-length fBm samples, and validated also on 1600-length fBm samples}
    \label{fig:benchm}
\end{figure}

In the first instance, we analyze the evaluation of the estimator on samples of the same length, i.e., of 1600, as the training was conducted. The visual representation of the results is displayed in Figure \ref{fig:benchm}. 
The upper right panel shows the estimated values versus those set for generating the trajectories. The overall performance of deep learning seems quite appealing in the graph. While the Hurst exponent values were chosen uniformly in the generation, the estimated values slightly deviate from that, as seen in the upper mid-panel histogram. The deviation indicates the presence of a bias at large -- close to one -- Hurst parameter values. With a few exceptions, the raw errors seem to follow normal law, as illustrated by the normal plot in the upper right corner.

Turning to absolute errors, its growth with the true Hurst value, as the lower left plot shows, is noteworthy. The histogram of the absolute errors in the middle of the lower row shows the skewed distribution close to that of an absolute value of a Gaussian variable. Remark, however, that two-sided deviations from the normal law add up, and the skewness of the absolute errors is 1.16, while the absolute value of a Gaussian variable has a skewness of 1. The maximum absolute error value is almost 0.06, more than four times the raw error's standard deviation (0.0147). That is regarded as extreme under the Gaussian law. It hints at a heavier-than-normal tail of the error distribution. 
The lower right plot exhibits the relative error. Understandably, when the 'true' generated Hurst value is extremely low, it induces high relative error even from a small absolute error. For this reason, we cut the graph and do not display the relative error when the generated Hurst value is lower than 0.02; such low values are not plausible in real applications anyway. As can be seen, the relative error depends heavily on the true value of the Hurst parameter. For that reason, together with the overall 95\% quantiles, we also calculate localized values. To a given generated Hurst value $H$, we take the $H-0.1, H+0.1$ interval around it and calculate the relative error quantiles from generated trajectories with true Hurst value from that interval. We present these quantiles, calculated in 0.1 steps, by the red line in the lower right plot.

We may change the loss function in the training from MSE to MAE, pushing the optimization to the conditional median. Not knowing what feature of the samples LSTM learns, it is difficult to understand the difference between the two optimizations. However, the difference in errors is really minuscule: RMSE: 0.014774 vs. 0.014866, MAE: 0.011582 vs. 0.011659 MRE\% 3.694370 vs. 3.631601 the first two to the advantage of the MAE, the last one to the MSE. Neither of the loss functions can be preferred, and the result indicates that well-balanced samples are generated for training. 

\begin{figure}
    \centering
    \includegraphics[width=0.90\textwidth]{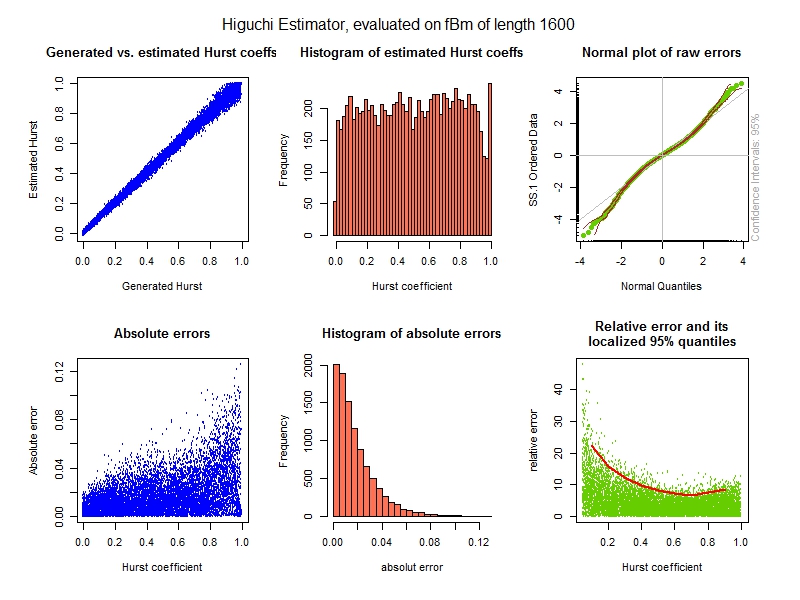}
    \caption{Caption}
    \label{fig:higu}
\end{figure}

For comparison, we present a similar analysis with the Higuchi estimator, noting that very similar results -- with alternate asymmetries though -- can be obtained by the detrended fluctuation analysis and the Whittle estimators, and results are significantly worse for the variogram-based estimator, not to mention the classical R/S estimator. We evaluate the Higuchi estimator on 10000 simulated fBm trajectories of length 1600 with uniformly distributed random Hurst parameters. The plot of estimated versus "true" generated values is shown in the upper left corner of Figure \ref{fig:higu}. Bias is now observable for the smallest (close to 0) and the largest (close to 1) values, as the histogram in the middle of the upper row discloses. According to the normal plot in the upper left corner, the estimator does not have a normal distribution; it has heavier than normal tails.

 The absolute errors slightly exceed that of the LSTM estimator and have heavier tails; this is how the first lower left corner plot and the histogram next to it show. The relative errors are not much higher than the ones for the LSTM estimator, though. That can be observed in the lower right plot. The quantiles are also similar in the low end of the Hurst parameters; however, they grow higher than that of the LSTM at high Hurst values.    \\

 Evaluating on fOU processes the LSTM estimator trained on fBm, we proceed similarly to the fBm case. The results are also very similar to the fBm case concerning both the LSTM and the Higuchi estimators. Therefore, we do not present a similar figure and detailed assessment separately; it would simply repeat what has been presented so far. In the first two blocks of rows in Table \ref{table:performance_metrics}, we exhibit the performance indicators of the LSTM network trained on fBm samples of length 1600 and evaluated on 10000 shorter (length 400), equally long, and longer (length 6400) samples of both fBm and fOU. The accuracy grows significantly by increasing the sample length; the errors almost halve by every means when quadrupling the sequence length.\\

 We also evaluate the LSTM on fOU processes when trained on 1600-long fOU processes. The generation of fOU processes is slower than fBm-s; hence, we are forced to contend with shorter epochs in training, consisting of 10000 trajectories only. The third block of rows in Table \ref{table:performance_metrics} reveals that the network's performance is almost as good as the one trained on fBm despite the shorter training on fOU. Compared to the second block of rows, we may notice that the evaluation is slightly better on shorter (400) samples, but on longer (6400) samples, its performance does not improve as much as the fBm-trained network. When the training and the evaluating sample lengths are equal, the two training types perform almost equally well. The higher maximum and quantiles of absolute errors in the 6400-long evaluation are due to several outliers. We guess that those outliers are the result of the shorter training. The skewness of the absolute errors is 2.3558, highly different from 1, the skewness of the absolute value of a normal distribution. It is unprecedentedly high, considering all of our experiments.

\begin{table}[h]
    \centering
    \begin{tabular}{ccccccccccccc}
        \toprule
        Type & Evaluation & RMSE & MAE & MRE\% & \multicolumn{3}{c}{Absolute Error} & \multicolumn{2}{c}{Relative Error \%} \\
         &length& & & & max & q95\% & q50\% & q95\% & q50\% \\
        \midrule
        LSTM & $400$ & $0.0311$ & $0.0241$ & $7.32$ & $0.1482$ & $0.0630$ & $0.0194$ & $24.45$ & $4.66$ \\
        trained on fBm 1600& $1600$ & $0.0149$ & $0.0117$ & $3.63$ & $0.0592$ & $0.0300$ & $0.0094$ & $11.73$ & $2.21$ \\
        and evaluated on fBm & $6400$ & $0.0079$ & $0.0066$ & $2.43$ & $0.0389$ & $0.0165$ & $0.0056$ & $9.80$ & $1.24$ \\
         \midrule
        LSTM & $400$ & $0.0307$ & $0.0238$ & $7.06$ & $0.1279$ & $0.0620$ & $0.0190$ & $22.59$ & $4.56$ \\
        trained on fBm 1600& $1600$ & $0.0156$ & $0.0122$ & $3.69$ & $0.0710$ & $0.0316$ & $0.0097$ & $11.95$ & $2.34$ \\
        evaluated on fOU& $6400$ & $0.0079$ & $0.0062$ & $2.02$ & $0.0293$ & $0.0157$ & $0.0051$ & $6.54$ & $1.20$ \\
         \midrule
        LSTM & $400$ & $0.0295$ & $0.0229$ & $7.02$ & $0.1270$ & $0.0599$ & $0.0184$ & $22.53$ & $4.33$ \\
        trained on fOU 1600& $1600$ & $0.0157$ & $0.0122$ & $3.62$ & $0.0708$ & $0.0321$ & $0.0098$ & $11.51$ & $2.31$ \\
        evaluated on fOU& $6400$ & $0.0107$ & $0.0085$ & $2.28$ & $0.1169$ & $0.0233$ & $0.0063$ & $6.65$ & $1.60$ \\
         \midrule
        Higuchi Estimator & $400$ & $0.0445$ & $0.0342$ & $10.5$ & $0.2507$ & $0.0903$ & $0.0268$ & $33.69$ & $6.57$ \\
        on fBm samples& $1600$ & $0.0246$ & $0.0182$ & $4.93$ & $0.1257$ & $0.0515$ & $0.0136$ & $14.34$ & $3.40$ \\
         & $6400$ & $0.0152$ & $0.0104$ & $2.61$ & $0.0895$ & $0.0332$ & $0.0071$ & $7.51$ & $1.84$ \\
        \bottomrule
    \end{tabular}
        \caption{Performance metrics for LSTM and Higuchi Estimator evaluated on fBm and fOU. The LSTM network was trained on 1600-long fBm samples.}
    \label{table:performance_metrics}
\end{table}
In summary, the LSTM estimator of the Hurst coefficients trained on fBm samples performs better than the traditional statistical estimators in every aspect. True, the LSTM halves the root mean squared errors compared to statistical estimators; however, it cuts the MAE to 60-70\% and the MRE  to 75-90\% only, depending on sequence length. LSTM also performs well on the maxima and the quantiles of the absolute errors; it almost halves those. However, RMSE and MAE are not sensitive to which range of the original parameter the errors occur. That points out the relevance of the relative errors, and in that terms, the performance of LSTM is not that excellent. In particular, when the 95\% quantiles of localized relative errors are considered, in the low range of the Hurst parameter, LSTM, and the statistical estimators are almost equally bad, nearing or even exceeding 20\%. On the high end of the parameter, nearing one, however, LSTM again significantly outperforms the Higuchi and other statistical estimators, as the latter's relative errors start to grow even along the Hurst values.\\

\subsection{Evaluation on Linear Fractional $\alpha$-Stable Motions}
In view of the explicit solution formula of the Langevin equation, the fOU process can be represented as a time-changed and scaled fBm, so it is not very surprising that the network trained on fBm samples performs well on fOU samples, too. Therefore, it is quite exciting to overstep this class and analyze the performance of LSTM on lfsm process samples. 

\begin{figure}
    \centering
    \includegraphics[width=0.90\textwidth]{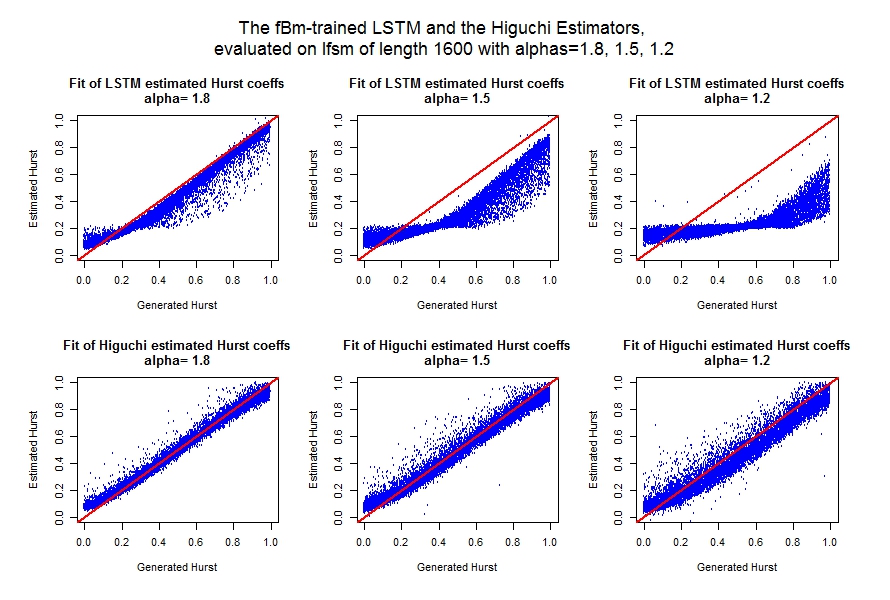}
    \caption{Hurst exponent estimation of lfsm-s of $\alpha=1.8,\ 1.5,\ 1.2$ by LSTM and Higuch methods and its errors. The network is trained on 1600-length fBm samples. The evaluation length is also 1600.}
    \label{figure:levydiag}
\end{figure}

As it has been said before, we cannot train the network on lfsm samples due to the lack of a fast generator at the moment. This is why we use the network trained on fBm and fOU samples. For the evaluatory analysis, we chose our benchmark, the LSTM network taught on 1600-long fBm samples in the span of 100 epochs. The evaluatory samples are generated by the rlfst package of R. We choose $\alpha$ from the range $(1,2)$ with the values 1.8, 1.5, and 1.2. With these parameters, the lfsm features long-range dependence. As Figure \ref{figure:levydiag} demonstrates, the LSTM network cannot properly detect the true Hurst parameter value, and the higher the index of stability, the more LSTM underestimates it. To the contrary, the Higuchi method estimates the Hurst exponent correctly, albeit with higher variance than in the case of fBm or fOU.

Even though the result concerning LSTM is negative, it is still valuable, as it points out the limitations of the described setup and training when evaluating the fBm-trained network on lfsm samples.

 
\subsection{The effects of training cycles and training sequence length}\label{effect}
In what follows, we consider the estimation quality depending on the training and evaluation length of the series. To better understand the sample-length dependence on the quality of learning, we refined the resolution. While keeping the methodology unchanged, we included further series of lengths 800 and 3200 in the training and 100, 200, 800, and 3200 long series in the evaluation. As it is partially indicative of the other error types, we restrict the evaluation to the RMSE only. The resulting errors are displayed in Table \ref{table:root_consistency}.

\begin{table}[h]
    \centering
    \begin{tabular}{@{}cccccccc@{}}%
        \toprule
         & \multicolumn{7}{c}{Root $MSE$ losses by evaluation seq. length}  \\
        \cmidrule(r{4pt}){2-8}
        training \\seq. length & $100$ & $200$ & $400$ & $800$ & $1600$ & $3200$ & $6400$\\ 
        \midrule
        $400$ & $0.0691$ & $0.0449$ & $0.0307$ & $0.0218$ & $0.0168$ & $0.0140$ & $0.0127$\\
        $800$ & $0.0693$ & $0.0447$ & $0.0302$ & $0.0210$ & $0.0152$ & $0.0116$ & $0.0094$\\
        $1600$ & $0.0708$ & $0.0459$ & $0.0311$ & $0.0211$ & $0.0149$ & $0.0106$ & $0.0079$\\
        $3200$ & $0.0738$ & $0.0472$ & $0.0312$ & $0.0213$ & $0.0149$ & $0.0105$ & $0.0080$\\
        $6400$ & $0.0748$ & $0.0480$ & $0.0318$ & $0.0217$ & $0.0151$ & $0.0110$ & $0.0083$\\
        \bottomrule
    \end{tabular}
    \caption{Root MSE losses of LSTM-based models trained on different sequence lengths.}
    \label{table:root_consistency}
\end{table}

Understandably, the performance is very weak when evaluated on the short series. Generally speaking, when \textit{trained} on shorter sequences, the performance of LSTM improves when \textit{tested} on longer sequences but is slower than LSTM variants \textit{trained} on longer sequences. LSTM variants trained on longer sequences still performed well on shorter sequences but not as well as dedicated variants. The differences in these cases are marginal, except for the very short series. The LSTM trained on 1600-long samples has the best performance, hardly distinguishable from the LSTM of 3200-long samples. However, training time is considerably less for the 1600-long, which is why we have chosen that setup as the reference. Training on 6400-long sequences does not improve the RMSE loss; the network probably reached its limit of learning, given the current architecture.

\begin{figure}
    \centering
    \includegraphics[width=0.65\textwidth]{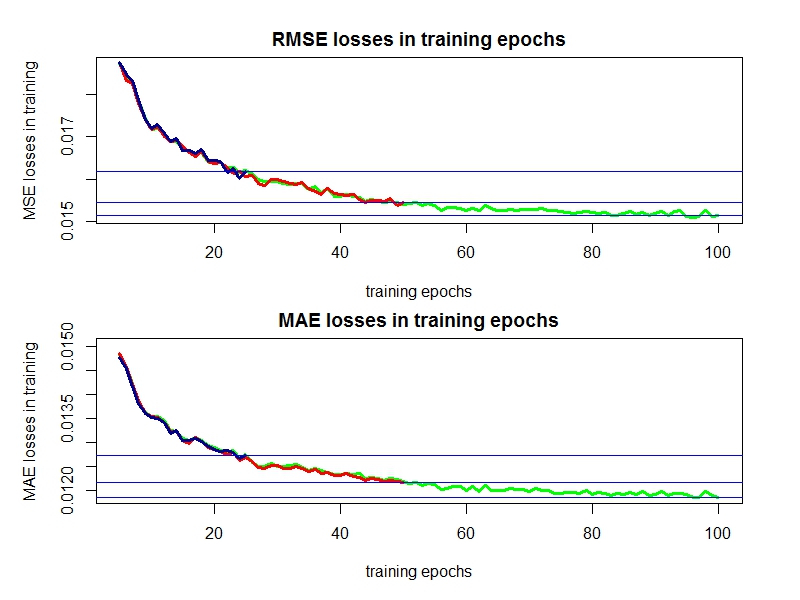}
    \caption{The changing of RMSE and MAE losses as training progresses. 25 epoch long training=blue, 50 epoch long training=coral, 100 epoch long training=green}
    \label{figure:loss}
\end{figure}


We analyzed the 1600-long fBm trajectory training, studying the loss dependence on training length in terms of the spanned epochs. Figure \ref{figure:loss} exhibits the loss changes during two series of 100, 50, and 25-long training, with MSE and MAE loss function choices, respectively. Although MSE is the loss function in the algorithm, we display the RMSE instead of the MSE in the figure and use it in the comparisons because of the very small MSE values. The graphs start from the 6th epoch. In the first 5 epochs, the loss is still so high that it would suppress the visibility of the changes in the remaining training epochs; hence, we omitted to display it. It is remarkable that the losses closely follow each other in independently started network teachings. Shorter teaching losses simply follow the beginning part of longer teaching. By the 50th training, the RMSE decreases to 95.1\% that of the 25th training, and by the 100th training, it decreases significantly less, to 98.6\% that of the 50th training. At the 75th step, the RMSE is only 0.3\% higher than the final RMSE loss. The same concerns the MAE losses in the respective teachings. The actual MAE loss percentages are: 50th/25th=95.5, 100th/50th=97.5\%. At the 80th step, the MAE is only 0.5\% higher than the final MAE loss. So, we may say that when using 1600-long fBm samples, there is no practical indication to increase the teaching epoch numbers above the 80th. Nevertheless, we used the 100 epochs in our analysis as the longest (and benchmark) teaching.

\section{An Application to Li-ion Battery Degradation}
In our study, we utilized a dataset provided by the NASA Prognostics Center of Excellence (PCoE), which includes experiments on Li-Ion batteries involving charging and discharging at different temperatures and recording the impedance as the damage criterion. The dataset is available at \url{https://www.nasa.gov/intelligent-systems-division/discovery-and-systems-health/pcoe/pcoe-data-set-repository/}.

\begin{figure}
    \centering
    \includegraphics[width=0.50\textwidth]{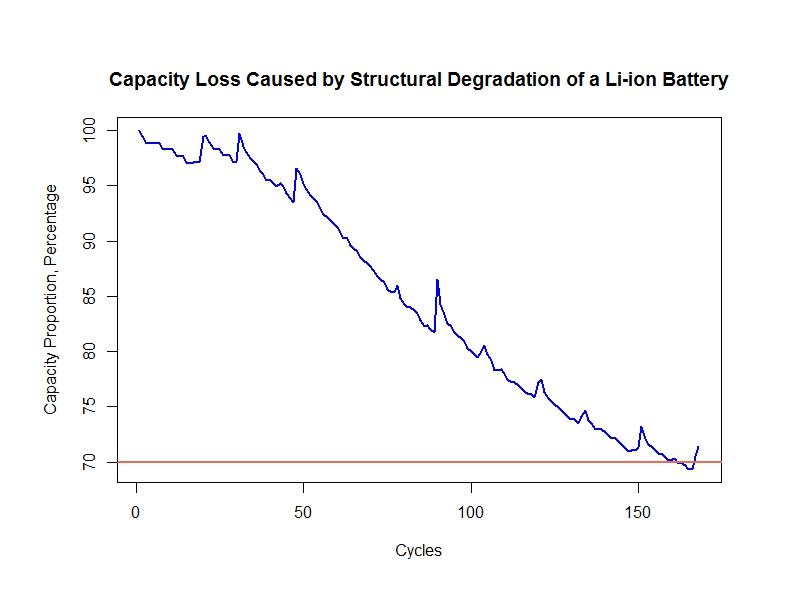}
    \caption{Capacity losses caused by structural degradation of Li-ion batteries}
    \label{figure:batt}
\end{figure}

We determined the capacity associated with the cycles, applying a 70\% threshold as the criterion for battery degradation. We present the obtained series in Figure \ref{figure:batt}. Our objective was to estimate the Hurst parameter of the process, assuming it follows fractional Brownian motion, based on the obtained data series.

We use the madogram to estimate the fractal dimension $D$, returning a Hurst exponent estimate by $2-D$, and the Higuchi method for statistical benchmarking. The madogram yielded a value of 0.8558, while the Higuchi method resulted in 0.6189.

In contrast, among our trained LSTM models, the best results from models trained on fBm samples were obtained from the 100- and 200-long training length, which estimated Hurst parameter values of 0.9312 and 0.9641, respectively. In subsection \ref{effect}, we concluded that the best estimate can be achieved when the network is taught on the same length series as the sample, for which to estimate the Hurst parameter. It may explain why the short training lengths provide the best results in the sense of being closest to the benchmark. The closest result to the madogram estimation was obtained from a model trained on a 200-length fractional Ornstein-Uhlenbeck process, with a value of 0.8294, and the 400-length fOU-trained network provided 0.7726 as the estimated Hurst parameter. The advantage of our approach is that we can also come up with confidence intervals in terms of absolute deviation and relative error. For the 100-long fBm training, the 95\% quantile of the absolute errors is 0.1280, whereas for the 200-long fBm training, the same 95\% quantile is 0.0875. Considering that the Hurst parameter is less than one, these quantiles produce 95\%  confidence intervals $(0.8032,1)$ and $(0.8361,1)$ for the two estimations, respectively, both containing the madogram-based estimation. Turning to the fOU-trained LSTM estimates, the absolute error's 95\% quantile is 0.0854 for the 200-long and 0.0612 for the 400-long training. These quantiles result in the 95\% confidence intervals $(0.744,0.9148)$ and $(0.7114,0.8338)$ for the corresponding estimates, respectively. Considering relative errors, the 95\% quantiles are far too large, being around 25\%. The exception is the case of the 400-long fOU trained network when the 95\% relative error quantile \textit{localized} around the 0.8 Hurst value is 8.78\%. It results in a confidence interval of $(0.7047,0.8404)$. Summarizing, we may say that in all models, Hurst values between 0.8 and 0.83 belong to the 95\% confidence interval.

In Shao and Si \cite{Shao2023}, several results obtained values between 0.8 and 0.9 for the Hurst parameter estimation using their model, which was defined as \(dX(t) = \mu dt + \sigma B_H(t)\), where \(X(t)\) represents the degradation process. This way, they assume a linear trend in the process. When modeling by fOU, as we did by applying the fOU-trained network, we introduce a trend of exponential (and random) characters that may play a similar role to the linear one. That may explain why ours and their estimations of the Hurst parameter are close to each other. Remark for clarity that the fOU used in the modeling is not stationary.

In Liu, Song, and Zio \cite{Liu2023}, an lfsm, called by them as fractional Lévy stable motion, was applied for modeling the degradation of lithium-ion batteries, where the Hurst parameter estimation yielded values around 0.6, aligning closely with our Higuchi estimation. The \(\alpha\) parameter value in their modeling was consistently around 1.8-1.9. Given that the LSTM estimator does not perform satisfactorily in the lfsm case, there is no controversy in our and their findings. Note the well-known fact that when \(\alpha=2\), the lsfm process reduces to a fractional Brownian motion.

Both articles focused on modeling the degradation of lithium-ion batteries, similar to our study, providing valuable insights and benchmarks for our analysis.

\section{Conclusions}

Our research demonstrates that a neural network with a standard architecture, when trained with a substantial volume of data, significantly outperforms traditional statistical estimators in both accuracy and speed for estimating process parameters, provided it is trained on appropriate process types. However, skewness in the data can adversely affect the accuracy of the deep learning estimator, particularly in scenarios involving rare but significant losses or risks, as highlighted by the error quantiles. Despite the overall high performance, the relative errors in the estimations can still be considerable under certain parameter configurations.

The training phase of the neural network may extend to several hours, but the estimation time remains significantly lower – by one or two orders of magnitude – compared to conventional methods. In terms of cross-process applicability, the network trained on fractional Brownian motion data also shows commendable performance when applied to fractional Ornstein-Uhlenbeck processes, but its effectiveness is reduced for fractional Lévy stable motions.

The methodology is applied to estimating the Hurst parameter in Li-ion battery degradation data. The obtained confidence bounds conform with the findings of previously published research.

Our results suggest that transforming the evaluation data can lead to impractical slowdowns in the estimation process, indicating that approaches like signature transform or certain transformer classes may not be promising alternatives. However, the current introduction of the iTransformer\cite{iTransformer2023} offers a promising alternative. The iTransformer inverts the duties of the attention mechanism and the feed-forward network, focusing on embedding individual time series as variate tokens to capture multivariate correlations effectively. The iTransformer's ability to handle multivariate correlations makes it a highly viable and efficient backbone for time series forecasting, addressing the limitations of traditional Transformer-based forecasters. Therefore, we intend to exploit its novel capabilities in our further research.

\section{Acknowledgement}
The last author gratefully acknowledges partial support from the Fulbright Teaching and Research Award while starting this research.

\end{document}